# AUTOMATIC AMBIGUITY DETECTION

*Richard Sproat, Jan van Santen*

Bell Labs – Lucent Technologies

## ABSTRACT

Most work on sense disambiguation presumes that one knows beforehand — e.g. from a thesaurus — a set of polysemous terms. But published lists invariably give only partial coverage. For example, the English word *tan* has several obvious senses, but one may overlook the abbreviation for *tangent*. In this paper, we present an algorithm for identifying interesting polysemous terms and measuring their degree of polysemy, given an unlabeled corpus. The algorithm involves: (i) collecting all terms within a $k$-term window of the target term; (ii) computing the inter-term distances of the contextual terms, and reducing the multi-dimensional distance space to two dimensions using standard methods; (iii) converting the two-dimensional representation into radial coordinates and using isotonic/antitonic regression to compute the degree to which the distribution deviates from a single-peak model. The amount of deviation is the proposed *polysemy* index.

## 1. INTRODUCTION

Published work on sense disambiguation (e.g. [12, 7, 2, 3, 4, 5, 8, 10]) invariably presumes that one has in mind a particular set of polysemous terms to be disambiguated. While this is quite often a reasonable assumption, it must be borne in mind that published lists invariably give only partial coverage. For example, the English word *tan* has several obvious senses — *WordNet*[1] lists two for the nominal interpretation, for instance — but one may overlook its function as an abbreviation for *tangent*.[2] It is desirable, therefore to have a method for detecting terms that display potential ambiguity in a language, which one might not necessarily expect to find listed in online lexical resources for that language. This is of particular interest for languages for which such on-line lexical resources do not exist, which includes most languages besides English and a few of the better studied European and Asian languages.

In this paper we present an algorithm for identifying interesting polysemous terms and measuring their degree of polysemy, given an unlabeled corpus. In addition to using standard techniques for dimensionality reduction, the algorithm also presents a novel application of *isotonic/antitonic regression*. We present and illustrate the algorithm in Section 2., and discuss some potential applications and limitations in Section 3.

The algorithm that we will present is most similar to previous work of Schütze [7, 8] in that the senses that the method "discovers" are arrived at by computing similarities among words that occur in contexts surrounding the target word. However, our approach differs from Schütze's primarily in proposing a *polysemy index*, that is a quantitative measure of how polysemous a word is, given the corpus. Thus given a list of, say, a few thousand target words, it is possible to rank them by their polysemy indices, with the most polysemous words (relative to the corpus used) appearing at the top of this list.

| | |
|---|---|
| **bass** | reed DOUBLE species instrument heard music 250 drums normal bows nebulifer Atlantic hunting ARTHUR secular–was continuo bays CAMPBELL CRAPPIE organ |
| **oxidation** | colorless microorganisms produces peroxide electrolysis nitrogen Considerable investigated dye conductor synthetic behind hydrochloric kiln hydrate bond smithsonite crystalline spontaneous combining |

**Table 1:** Twenty randomly selected words that cooccur with *bass* and *oxidation* in the 10 million word *Grolier's Encyclopedia*. Parameters: $k = 30$, $\mathrm{threshold} = 0.005$, $\mathrm{mincount} = 1$. Words are capitalized as they appear in the corpus.

## 2. THE ALGORITHM

1. For target term $w$, collect all terms q occurring within a $k$-term window (typically $k = 30$) of all instances of $w$ in corpus $C$. $FREQ_w(q)$ is the frequency of $q$ within such $k$-term windows, and $FREQ_C(q)$ is the frequency of $q$ in $C$. $REL_w$ is the set of $n_w$ terms $q$ such that

   $$FREQ_w(q)/FREQ_C(q) > \mathrm{threshold} \text{ (typically } \mathrm{threshold} = 0.005)$$

   and

   $$FREQ_w(q) > \mathrm{mincount} \text{ (typically } \mathrm{mincount} = 1)$$

   Table 1 gives a randomly selected selection of words that cooccur with the words *bass* and *oxidation*, two nouns that occur with roughly the same frequency in the 10 million word *Grolier's Encyclopedia*.

2. For each $q, q'$ in $REL_w$, $CO_w(q'|q)$ denotes the frequency of $q'$ within a $W$-term window ($W$ typically 100) of (all instances of) $q$. Define a symmetric $n_w \times n_w$ distance matrix $D_w$, with elements:

   $$DIST_w(q,q') = 1 - \frac{\left[\frac{CO_w(q|q')}{FREQ_C(q)} + \frac{CO_w(q'|q)}{FREQ_C(q')}\right]}{2}$$

[1] See `http://www.cogsci.princeton.edu/~wn/`, and [5] for further references.

[2] This particular ambiguity is important for a text-to-speech application since it involves a pronunciation difference, though of course most ambiguities do not involve a pronunciation difference.

Using methods from [9], the terms in $REL_w$ can be represented 2-dimensionally such that distances in the representation correlate maximally with distances in $D_w$. For strongly polysemous terms, the representation typically shows elongated, near-linear clusters — one cluster per sense — that radiate outwards from a common region. For less polysemous terms, such radial structure is lacking.

Figure 1 shows the first two dimensions for *bass* and *oxidation* as computed over *Grolier's Encyclopedia*.

3. For each $q$, compute $\text{distance}_w(q)$ from the origin in the 2-dimensional representation (the common region is near the origin) and $\text{cosine}_w(q)$ with respect to the horizontal axis. Given $\text{cosine}_w(q)$, select one of 90 4-degree radial bins, and increment the entry for that bin by $\text{distance}_w(q)$. After smoothing, polysemous terms typically have multiple widely-separated regions with high entry values, yielding a multi-peaked plot. For relatively non-polysemous terms, a single, broad peak will be observed. We apply isotonic/antitonic regression [1] forcing a single-peaked fit, and propose the inverse goodness-of-fit as an overall *polysemy index*. Briefly, this method works as follows. We start with the isotonic regression for a sequence of numbers $x_1, \cdots, x_n$, which is defined to be a sequence $\hat{x}_1, \cdots, \hat{x}_n$ that minimizes the sum of squared differences between $\hat{x}_i$ and $x_i$ subject to the constraint that $\hat{x}_1 \leq \hat{x}_2 \leq \cdots \leq \hat{x}_n$; for anti-tonic regression, we replace "$\leq$" by "$\geq$". In other words, the isotonic (anti-tonic) regression is the best fitting non-decreasing (non-increasing) curve. To measure single-peakedness, we compute for each point $i$ the isotonic regression for the points $1, \cdots, i$ and the anti-tonic regression for the points $i, \cdots, n$, and add the two sums of squared differences. We then define the peak to be that value of $i$ for which this combined sum is minimized. The value of this minimum is a measure of the degree to which the points can be fitted with a single-peaked function. The strength of this method is that no assumptions whatsoever are made about the shape of the single-peaked function.

   Figure 2 shows the smoothed distance-versus-radial-bin plots for *bass* and *oxidation* as computed over *Grolier's Encyclopedia*.

As an instance of the output of the algorithm, consider the terms with the twenty-five highest polysemy indices, along with their "discovered" senses, derived from a list of about 9,000 lower case terms ($50 < FREQ(w) < 5000$) in Grolier's Encyclopedia. These data are shown in Table 2. Also listed, for each term, is whether or not the discovered sense ambiguity is found in *WordNet*. Clearly most of the discovered senses (as well as other senses not discovered by the algorithm in this corpus) are to be found in *WordNet* — though there are some omissions, such as the discovered senses of *inherited, aromatic* and *r*. But note again that *WordNet* and *WordNet*-like resources exist for only a few languages meaning that for other languages a tool such as the one described here could potentially be useful. Also, to the extent that one finds senses that have been overlooked by the creators of on-line thesauruses, such a tool can also be useful for English.[3]

| term | senses | index | *wn* |
|---|---|---|---|
| bill | beak; legislation | 40.08 | Y |
| mild | vs. intense; weather | 32.39 | Y |
| seasons | weather; performance | 31.69 | Y |
| inherited | genetics; dynasties | 29.44 | N |
| moderate | weather; use of alcohol | 28.60 | N |
| garden | plot of ground; plants | 28.35 | Y |
| gardens | plot of ground; plants | 26.69 | Y |
| bass | music; fish | 26.61 | Y |
| tip | tip of peninsula; (sharp) point | 24.98 | Y |
| bills | beak; legislation | 24.21 | Y |
| mate | biology; "running mate" | 22.67 | Y |
| aromatic | odor; chemistry | 22.25 | N |
| neutral | chemistry; alliances | 21.86 | Y |
| r | letter; abbr. for *reigned* | 21.31 | N |
| destruction | ?? | 20.84 | N |
| fur | hair; fur (commodity) | 20.71 | Y |
| docked | space ships; tails of dogs, etc. | 20.57 | Y |
| running | racing; candidacy | 19.92 | Y |
| M.D | ?? | 19.59 | N |
| paint | painting as art; wallcoverings | 19.37 | Y |
| conducting | physics; orchestral | 19.17 | Y |
| forestry | forestry service; field of study | 19.09 | N |
| platform | base; political platform | 19.06 | Y |
| fly | flying; diptera | 19.05 | Y |
| arm | geography; weapons | 18.94 | Y |

**Table 2:** Terms with the twenty-five highest polysemy indices, along with their discovered senses, derived from a list of about 9,000 lower-case terms ($50 < FREQ(w) < 5000$) in Grolier's Encyclopedia. These data are shown in Table 2. Parameter settings are as for Figures 1–2. Examples tagged with "??" are cases where the sense distinction, if there is one, is subtle from the point of view of human judgment of polysemy. The fourth column, labeled "*wn*" indicates whether the particular set of discovered senses is listed in *WordNet*.

## 3. DISCUSSION

Besides identifying interesting polysemous terms, the algorithm can also be used with known polysemous terms to select canonical contexts for the various senses: one selects contexts containing terms falling in bins near each peak. Such contexts can then be used to seed a self-organizing disambiguation method, such as that proposed by Yarowsky [12]. For instance, the following two sets of terms occur within plus or minus two bins of each of the two peaks for *bass*, as computed over *Grolier's* encyclopedia:

1. 1.8, geese, Resources, deer, grows, grouse, lb, chitin, Pacific, gravel, forage, females, eggs, ecological, white-tailed, coast, weigh, forked, 2.1, Suwannee, turkey, valuable, ducks, striped, nest, CAMPBELL, skunk, snake, spotted, sea, raccoon, quail, bituminous, carnivorous, birds, sand, species, Animal, hemlock, Mineral, streams, temperate, bays, lakes, laterally, kg, squirrel, lake, basin, bear,

[3] A note on speed. The algorithm as currently implemented is not speedy: the 9,000 terms just discussed took about 500 hours on an SGI Origin 200. However, a number of things could be done to speed the algorithm up.

cooler, cm, yellow, reach, reaches, relatives, reserves, transitional, stray, shoulder, reservoirs, pollution, giant, eastern, drainage, desirable, gray, limestone, insufficient, hunting, guards, 0.9, 45, Strait, Bass, North, Florida, Erie, male, currents, female, naturally, 550, Trout, heads, sparse, sport, uneven, compressed, Northern, hair, ground, U., reproduction, repeats, worldwide, Utah, exceeds, 150, prefer, reproduce, belongs, Along, robust, barred, ARTHUR, resources, sportsmen, 250, VOICE, families, differently, Texas, America, extend, Portugal, PATTON, Alabama, Michigan, white, Indigenous, Common, lengths, majority, midway, weights, native, westward, Dominions, Kentucky, restricted, leg, Eventually, TURNER, rare, rarely, descended, Usually, reinforced, 1500-1800, stream, generalized, Illinois, Massachusetts, slightly, Made, Probably, horns, Several, None, common, shorter

2. 1750, 17th, drum, theme, tuba, makers, scores, Gilmore, finger, score, song, lyric, treble, intervals, Fyodor, sound, piece, beat, played, accompany, neglected, reed, recorders, voice, shorthand, written, singer, valved, singers, soprano, baritone, brass, heard, loudspeakers, improvisation, organ, marching, contralto, octaves, notes, octave, playing, pitch, guitars, notation, saxophone, note, trombone, Vienna, drums, tone, singing, WIENANDT, Boris, jazz, textures, Wolfgang, tones, cymbals, Ludwig, tenor, OBOE, Donna, performance, BASS, Franz, vocal, ELWYN, baroque, woodwind, flutes, flute, tuned, chorus, chromatic, rhythm, viol, guitar, strings, ensemble, harmonic, fretted, double-reed, string, VIOLIN, symphony, trumpets, violas, violins, chamber, voices, harmony, bop, bugle, alto, accompaniment, d'amore, Wagner, trombones, concert, choir, clarinet, composing, compositions, E-flat, music, performances, performer, performers, instruments, solo, right-hand, melody, melodic, percussion, resonator, CLARINET, orchestras, lute, keyboard, musical, improvised, sounded, operatic, B-flat, harps, MUSIC, instrumental, instrument, Godunov, musicians, reggae, piccolo

It is important to understand a couple of weaknesses of the approach. First of all, the particular version of the method we have described is a "bag of words" model, since the only raw data that are used are the frequencies of occurrence of words within a given window. As such it shares with other similar approaches, such as Schütze's, the limitation that it works best for ambiguities that are topical; it is for such ambiguities that the occurrence of particular words in the context are most likely to serve as useful clues. Thus, ambiguities in nouns (which are often topical, though see [5, page 152] for some discussion of this point) are more easily detected than ambiguities in verbs: ambiguities in verbs tend to relate to much more restricted information, such as the particular head noun of the object NP used with the verb (cf. [6]). However there is no principled reason why the same algorithm could not also be used for such cases. All that would be required would be to provide a method $M$ for extracting the relevant contextual features (e.g., the head noun of the object NP of a verb). Then one could apply the algorithm in Section 2. as before, replacing the collection of "all terms q occurring within a $k$-term window of all instances of $w$ in corpus $C$" with $M$.

Second, the method works well — that is, the terms with the highest polysemy indices are words that have clearly distinct senses according to human intuitions — only when the text corpus comprises texts covering a wide range of topics; this is characteristic of encyclopedia text and explains why we have used *Grolier's* encyclopedia for our experiments. There is nothing unusual in this observation: in a similar vein, Yarowsky [11, 12] used Grolier's encyclopedia in his thesaurus-category-based experiments for precisely the same reasons of coverage. Of course, this does imply that in order to run the same method on another language besides English, we would first need to acquire an encyclopedia or encyclopedia-like corpus for that language, something that we have not currently done.

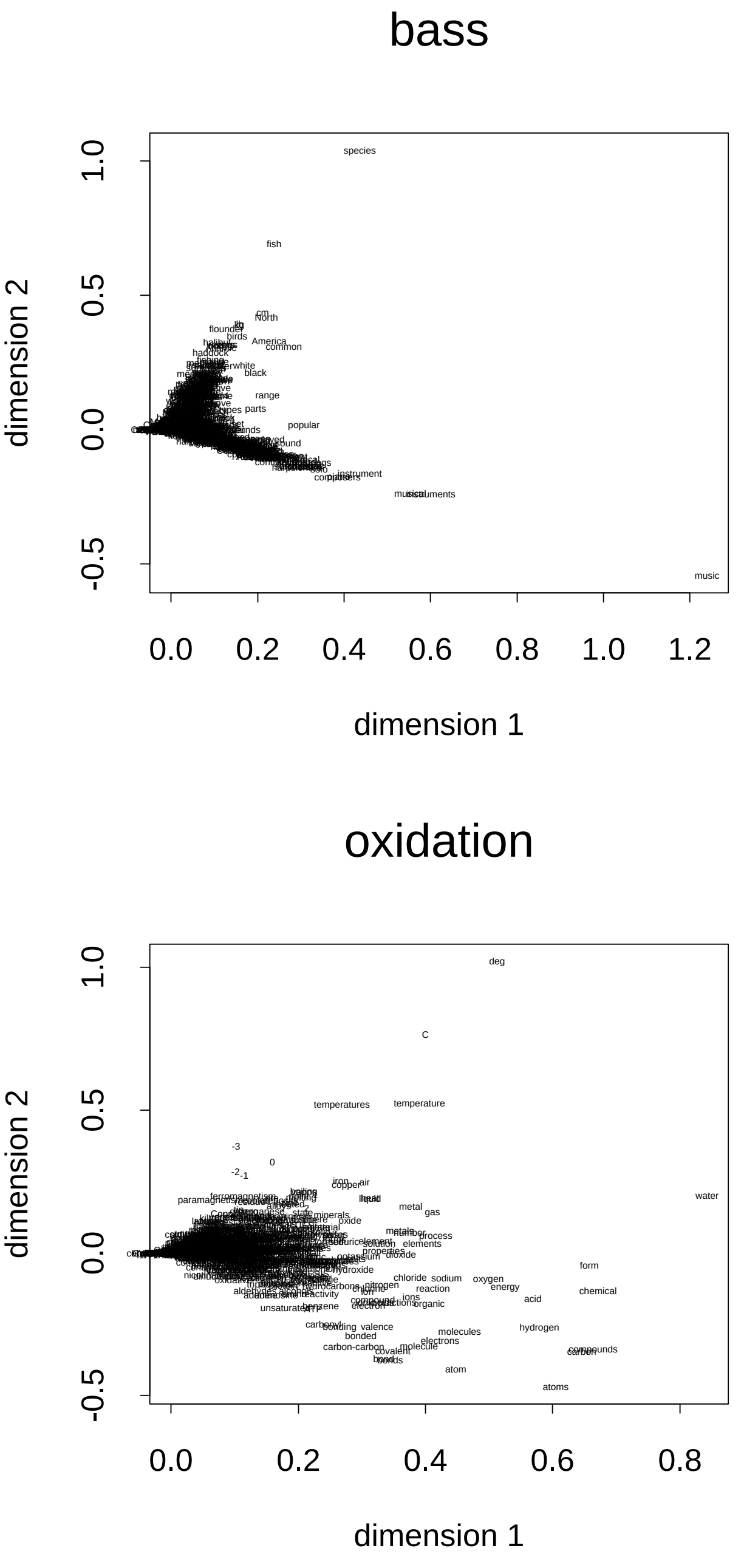

**Figure 1:** First two dimensions for *bass* (top panel) versus *oxidation* (bottom panel), as computed over *Grolier's Encyclopedia*, with the following parameter settings: $k = 30$, $\mathrm{threshold} = 0.005$, $\mathrm{mincount} = 1$, $W = 100$.

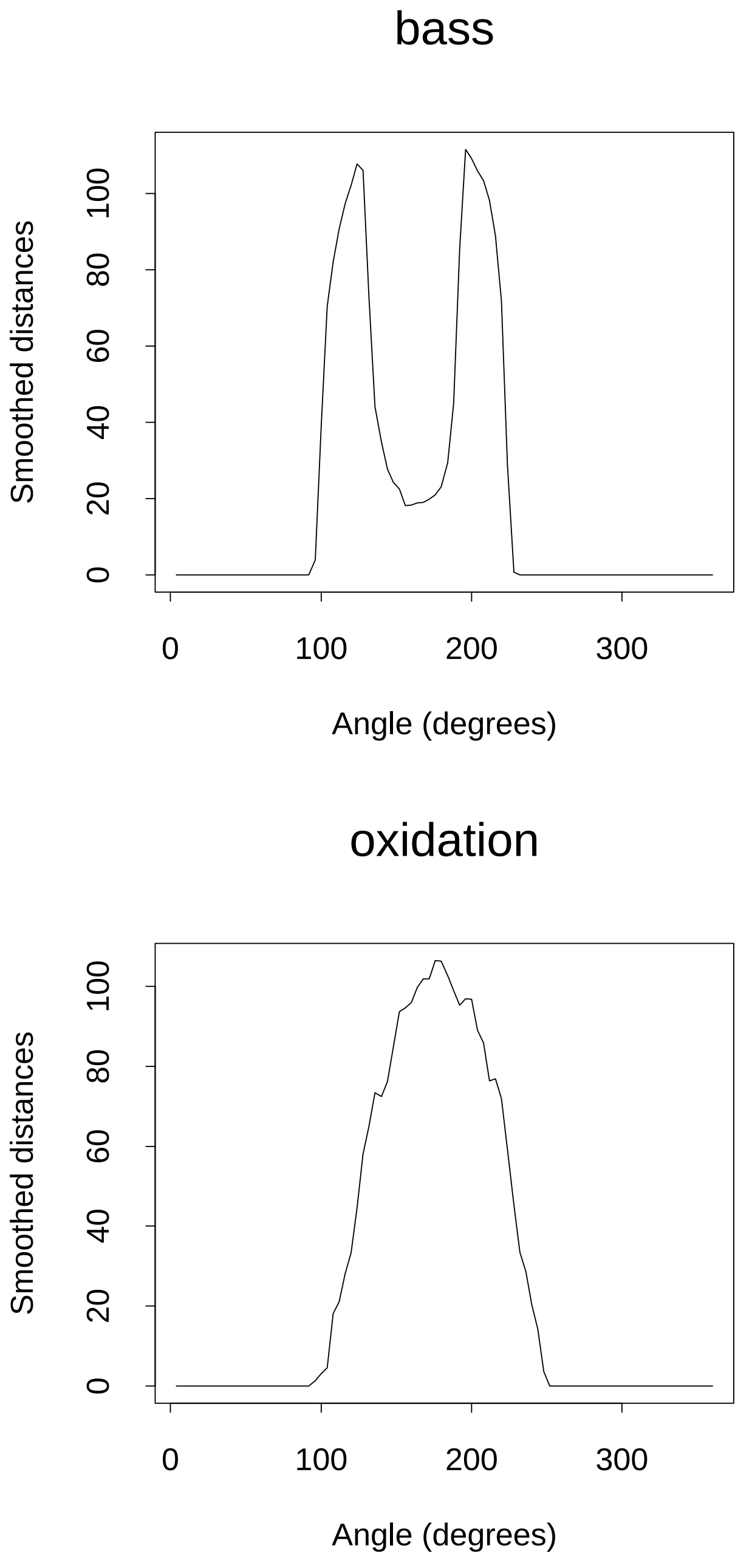

**Figure 2:** Smoothed distances (abscissa) plotted against radial bins (ordinate) over 360 degrees, for *bass* (top panel, $\mathrm{polysemy\ index} = 26.61$) versus *oxidation* (bottom panel, $\mathrm{polysemy\ index} = 0.23$), as computed over *Grolier's Encyclopedia*, with the following parameter settings: $k = 30$, $\mathrm{threshold} = 0.005$, $\mathrm{mincount} = 1$, $W = 100$.